\pdfoutput=1

\documentclass[11pt]{article}

\usepackage[preprint]{coling}

\usepackage{times}
\usepackage{latexsym}

\usepackage[T1]{fontenc}

\usepackage[utf8]{inputenc}

\usepackage{microtype}

\usepackage{inconsolata}

\usepackage{graphicx}

\usepackage{booktabs}
\usepackage{xcolor}
\usepackage{colortbl}
\usepackage{multirow}

\usepackage[T1]{fontenc}

\usepackage[utf8]{inputenc}

\usepackage{microtype}

\usepackage{inconsolata}

\usepackage{graphicx}

\title{Beyond Data Quantity:\\ Key Factors Driving Performance in Multilingual Language Models}

\author{Sina Bagheri Nezhad,  Ameeta Agrawal,  Rhitabrat Pokharel\\
        Department of Computer Science \\ Portland State University, USA \\ \{sina5, ameeta, pokharel\}@pdx.edu}

\begin{document}
\maketitle
\begin{abstract}
Multilingual language models (MLLMs) are crucial for handling text across various languages, yet they often show performance disparities due to differences in resource availability and linguistic characteristics. While the impact of pre-train data percentage and model size on performance is well-known, our study reveals additional critical factors that significantly influence MLLM effectiveness. Analyzing a wide range of features, including geographical, linguistic, and resource-related aspects, we focus on the SIB-200 dataset for classification and the Flores-200 dataset for machine translation, using regression models and SHAP values across 204 languages. Our findings identify token similarity and country similarity as pivotal factors, alongside pre-train data and model size, in enhancing model performance. Token similarity facilitates cross-lingual transfer, while country similarity highlights the importance of shared cultural and linguistic contexts. These insights offer valuable guidance for developing more equitable and effective multilingual language models, particularly for underrepresented languages.
\end{abstract}

\section{Introduction}

Multilingual language models have garnered significant attention due to their ability to handle and generate text across various languages, playing a crucial role in tasks such as machine translation, cross-lingual information retrieval, and multilingual content creation. However, achieving fair and effective performance across languages with diverse linguistic characteristics and varying resource availability remains a formidable challenge.

Prior research has identified several features that influence the performance of multilingual language models \cite{zhong2024englishcentricllmslanguagemultilingual,bagheri-nezhad-agrawal-2024-drives,zhu-etal-2024-multilingual,chau-smith-2021-specializing}. Although many factors are widely acknowledged to impact model performance, potentially even in a manner similar to the butterfly effect, these studies have often focused on a limited set of features. In contrast, our work aims to conduct a comprehensive analysis to systematically explore and quantify the effects of a broader range of features. Specifically, we examine 12 distinct features related to both the models and the languages they are designed to process.

\begin{figure*}
    \centering
    \includegraphics[width=1\linewidth]{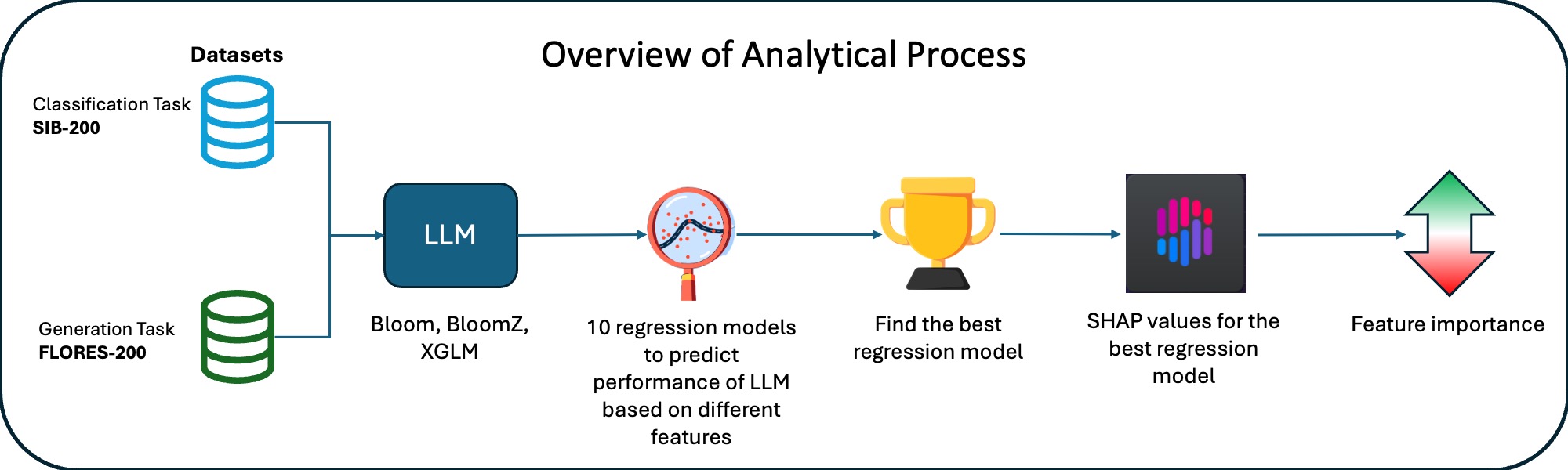}
    \caption{Overview of the Analytical Process to Determine Feature Importance on LLM Performance: Starting with datasets (SIB-200 for classification and FLORES-200 for generation), we applied various multilingual language models (LLMs) and evaluated their performance. Using regression models, we predicted LLM performance in different languages based on model and language features, selected the best-performing regression model, and analyzed it with SHAP values to identify feature importance.} 
    \label{fig:overview}
\end{figure*}

In this study, we analyze the performance of multilingual language models (Bloom, XGLM and BloomZ in different sizes) in 204 languages, using both classification (SIB-200 dataset \cite{adelani-etal-2024-sib}) and generation (Flores-200 dataset \cite{nllb2022flores}) tasks. We evaluate these models in zero-shot and two-shot learning settings, considering 14 different model configurations and sizes. Our experiments involve over 2.3 million instances, providing a robust basis for our analysis.\footnote{The code for this study is publicly available at \href{https://github.com/PortNLP/SHAP-MLLM-Analysis}{https://github.com/PortNLP/SHAP-MLLM-Analysis}.} Figure \ref{fig:overview} shows the overview of the analysis.

The primary contributions of this paper are as follows:

\begin{itemize}
    \item \textbf{Comprehensive Feature Analysis:} We investigate the impact of 12 distinct features, encompassing model-specific attributes (e.g., model size, pre-train data percentage) and language-specific attributes (e.g., script type, geographical proximity), to understand their influence on model performance across a diverse set of languages.
    \item \textbf{Evaluation Across Tasks and Configurations:} Our study spans both classification and generation tasks, assessed in zero-shot and two-shot learning settings. We consider multiple model architectures and sizes, offering insights into how different configurations affect multilingual model performance.
    \item \textbf{Quantitative Assessment of Feature Importance:} We employ SHAP (SHapley Additive exPlanations) values to quantify the importance of each feature \cite{NIPS2017_7062}, providing a detailed understanding of the factors driving performance disparities in multilingual language models.
    \item \textbf{Implications for Fair and Effective Multilingual Modeling:} Our findings offer practical guidance for developing more equitable and effective multilingual language models, particularly for underrepresented languages, by highlighting the features that most significantly impact model performance.
\end{itemize}


\section{Related Work}
The development and evaluation of multilingual language models have been widely studied, with models like mBERT, XLM-R, Bloom, XGLM, and Llama 3.1 demonstrating their capability to handle multiple languages with varying resource levels effectively \cite{devlin-etal-2019-bert, conneau-etal-2020-unsupervised, workshop2023Bloom176bparameteropenaccessmultilingual, lin-etal-2022-shot, dubey2024llama3herdmodels}. Despite these advancements, achieving fair performance across diverse languages remains challenging.

Recent efforts, such as the Glot500 project and the BigTranslate project, have focused on expanding multilingual corpora and enhancing translation capabilities, emphasizing the need for inclusive benchmarks and tailored training approaches \cite{imanigooghari-etal-2023-glot500, yang2023bigtranslateaugmentinglargelanguage}. Additionally, studies have explored key factors driving multilingual model performance, highlighting the importance of language-specific features and data distribution \cite{nezhad2024exploringmazemultilingualmodeling, bagheri-nezhad-agrawal-2024-drives}.

Tokenization is a critical aspect of multilingual modeling, where the choice of tokenizer and vocabulary allocation significantly impacts cross-lingual transfer and task performance \cite{pires-etal-2019-multilingual, wu-dredze-2019-beto, lample2019cross}. Successful cross-lingual transfer is influenced by shared vocabulary, linguistic similarity, and training data availability, as discussed in a comprehensive review by \citet{philippy-etal-2023-towards}.

Despite advancements in understanding multilingual language models, most studies focus on a narrow set of features or tasks. Our work fills this gap by analyzing 12 features across 204 languages, covering both classification and generation tasks in different learning settings. 

\section{Methodology}

In this section, we detail the datasets used, the models evaluated, the features extracted, and the evaluation methods employed in our study.

\subsection{Dataset Description}
We used two datasets in our experiments: SIB-200 for classification tasks and Flores-200 for generation tasks.

\paragraph{Flores-200 Dataset}
Flores-200 is a multi-way parallel corpus with sentences translated into over 200 languages, widely used to benchmark machine translation and multilingual models. It highlights performance gaps between high- and low-resource languages, promoting inclusive evaluations \cite{nllb2022flores}. The test set includes 204 languages, each with 204 instances.

\paragraph{SIB-200 Dataset}
SIB-200, based on Flores-200, is an open-source benchmark for topic classification across 200+ languages and dialects, addressing NLU dataset gaps for low-resource languages \cite{adelani-etal-2024-sib}. Its test set also covers 204 languages, with 204 instances per language.

\subsection{Model Configuration}
We conducted a direct evaluation of three multilingual models: Bloom, BloomZ, and XGLM, each tested across various sizes. Although newer multilingual models, such as Llama 3.1 \cite{dubey2024llama3herdmodels}, are now available, we selected these models because they were trained on a wide range of languages, are represented in different model sizes, and have accessible training dataset statistics. This makes them ideal for our comprehensive analysis of multilingual language model performance.

\textbf{Bloom} is a large language model developed by the BigScience collaboration, trained on the ROOTS corpus and capable of generating text in 46 natural languages and 13 programming languages. For our experiments, we used five sizes of Bloom, ranging from 560 million to 7.1 billion parameters \cite{workshop2023Bloom176bparameteropenaccessmultilingual}.

\textbf{BloomZ} is a fine-tuned variant of Bloom, optimized with multitask prompts to improve performance on specific tasks. We evaluated the same sizes as Bloom, ensuring consistency in comparisons \cite{muennighoff-etal-2023-crosslingual}.

\textbf{XGLM} is another multilingual model trained on 30 natural languages. The four sizes tested for XGLM ranged from 564 million to 7.5 billion parameters \cite{nllb2022flores}.

\subsection{Features}
We extracted a variety of features to analyze their impact on model performance. These features encompass geographical, linguistic, token similarity, and training-related aspects, including a total of 12 features drawn from both model characteristics and language-specific attributes.

\subsubsection{Model features}
In our analysis, we considered several key features related to the language models themselves, including model size, the distribution of pre-training data, and Instruction tuning data (specifically for BloomZ). 
\begin{enumerate}
    \item \textbf{Model size} refers to the number of parameters, impacting the model’s learning capacity. We examined models of various sizes to see how capacity affects multilingual performance.
    \item \textbf{Pre-training data} represents the language distribution in the initial training data, helping assess its impact on cross-language generalization.
    \item \textbf{Instruction tuning data} involves additional datasets for refining models on instruction-based tasks, particularly in BloomZ.
\end{enumerate}

\subsubsection{Language features}
To examine the impact of geography and culture on language models, we analyze two distinct features: geographical proximity and country similarity.
\begin{enumerate}
\setcounter{enumi}{3}
    \item \textbf{Geographical proximity} represents the physical distance between languages, derived from latitude and longitude data from Glottolog \cite{glottolog2024}. This feature, reduced with Multi-Dimensional Scaling (MDS) \cite{kruskal1964multidimensional}, captures linguistic traits influenced by regional contact, such as phonetic or lexical similarities arising from shared landscapes or historical migrations.
    \item \textbf{Country similarity}, in contrast, captures sociopolitical and cultural overlap by identifying the countries where each language is spoken (also sourced from Glottolog \cite{glottolog2024}). Using a Jaccard similarity matrix, reduced with MDS, this feature emphasizes shared cultural and linguistic traits, even among geographically distant languages that coexist within similar cultural or political spheres.
\end{enumerate}

Linguistic features were extracted by considering both the language family and the script used for each language. 
\begin{enumerate}
\setcounter{enumi}{5}
    \item \textbf{Language family} for each language was obtained from Ethnologue including their genetic classifications \cite{ethnologue2024}.
    \item \textbf{Script type} refers to the specific writing system used by a language, identified by ISO 15924 codes \cite{iso15924}, which categorize scripts based on their visual and structural characteristics. This information was directly available in the datasets we used.
\end{enumerate}

Both language family and script are categorical variables. To include these categorical variables in our regression models, we applied one-hot encoding.

Although script type is an important factor in our analysis, token similarity provides a more granular view of linguistic overlap at the lexical level, which is crucial for understanding how languages may influence one another in a multilingual model.

\begin{enumerate}
\setcounter{enumi}{7}
    \item \textbf{Token similarity}, measuring vocabulary overlap between languages, offers insight into linguistic similarity. We tokenized the SIB-200 train-set using model-specific tokenizers and calculated Jaccard similarity between token sets. This similarity matrix was then reduced to ten features using MDS.
\end{enumerate}

Additionally, we included Socio-Linguistic and Digital Support Features, which offer insights into the demographic, vitality, and digital presence of languages. These ordinal features -- population, language vitality, digital support, and resource level -- were numerically encoded to preserve their ordinal nature for regression analysis.

\begin{enumerate}
\setcounter{enumi}{8}
    \item \textbf{Population} data, sourced from Ethnologue, categorizes the number of speakers for each language into ranges like `10K to 1 million', `1 million to 1 billion', and `1 billion plus' \cite{ethnologue2024}.
    \item \textbf{Language Vitality} is categorized by Ethnologue into `Institutional', `Stable', `Endangered', and `Extinct', reflecting the language's community support and risk of endangerment or extinction \cite{ethnologue_vitality2024}.
    \item \textbf{Digital Language Support} assesses a language's digital presence, including content, localization tools, and resources. Ethnologue categorizes this support from `Still' (no digital presence) to `Thriving' (comprehensive digital ecosystem) \cite{ethnologue_digitalsupport2024}.
    \item \textbf{Resource Level} refers to the availability of linguistic resources like dictionaries and grammars for each language. \citet{joshi-etal-2020-state} classify languages into six levels, from those with minimal resources (Class 0) to those with extensive support (Class 5), reflecting varying levels of resource availability and digital advancement potential.
\end{enumerate}

\subsection{Feature Analysis}
To evaluate multilingual language model performance, we conducted a comprehensive analysis across classification and translation tasks, testing each of the 14 models in zero-shot and two-shot in-context learning settings \cite{NEURIPS2020_1457c0d6}. This dual-task evaluation enabled us to assess model performance across different languages and learning scenarios, providing insights into their effectiveness in handling multilingual data.

For the \textbf{classification task}, we used the SIB-200 dataset, calculating F1 scores based on model outputs compared to ground truth for each language.

For the \textbf{generation task}, we translated from various languages to English using the Flores-200 dataset, assessing accuracy with sacreBLEU scores against reference translations \cite{post-2018-call}.

To better understand the factors influencing model performance and to quantify the relationships between input features and performance metrics (F1 and sacreBLEU scores), we applied ten regression models: \textbf{Linear Regression} \cite{55e7ba22-38fb-3d2b-9a2c-0e68080abfc3}, \textbf{Random Forest} \cite{breiman2001random}, \textbf{Decision Tree} \cite{quinlan1986induction}, \textbf{Support Vector Regression (SVR)} \cite{Vapnik1995SupportVectorNetworks}, \textbf{Gradient Boosting} \cite{friedman2001greedy}, \textbf{XGBoost} \cite{chen2016xgboost}, \textbf{K-Nearest Neighbors} \cite{60c19788-1128-3b5f-9275-2d63cc155832}, \textbf{Lasso} \cite{51791361-8fe2-38d5-959f-ae8d048b490d}, \textbf{Ridge} \cite{doi:10.1080/00401706.1970.10488634}, and \textbf{Elastic Net} \cite{10.1111/j.1467-9868.2005.00503.x}.

We split the data into an 80-20 training-test split and assessed each model’s performance using R-squared ($R^2$) and Mean Squared Error (MSE), providing a robust evaluation of predictive accuracy across different language and model configurations.

To further understand the impact of each feature on model performance, we utilized SHAP (SHapley Additive exPlanations) values, which offer a unified measure of feature importance for each prediction \cite{NIPS2017_7062}. We focused on models that demonstrated strongest predictive performance for each task, and analyzed both individual and aggregated (abstract) features to gain insights into broader categories like geographical, linguistic, and token similarity. This analysis provided a deeper understanding of how these features contribute to overall model performance.

\begin{table*}[ht]
\centering
\begin{tabular}{lcccc}
\toprule
\textbf{Task} & \textbf{Setup} & \textbf{Bloom} & \textbf{BloomZ} & \textbf{XGLM} \\
\midrule
\multirow{4}{*}{Classification} 
 & \multirow{2}{*}{Zero-Shot} & Random Forest & Random Forest & XGBoost \\
 & & \small $R^2 = 0.645$, \small MSE = 0.005 & \small $R^2 = 0.903$, \small MSE = 0.001 & \small $R^2 = 0.855$, \small MSE = 0.003 \\
\cmidrule(lr){2-5}
 & \multirow{2}{*}{Two-Shot} & XGBoost & Gradient Boosting & XGBoost \\
 & & \small $R^2 = 0.847$, \small MSE = 0.007 & \small $R^2 = 0.754$, \small MSE = 0.009 & \small $R^2 = 0.902$, \small MSE = 0.003 \\
\midrule
\multirow{4}{*}{Generation} 
 & \multirow{2}{*}{Zero-Shot} & Gradient Boosting & Gradient Boosting & XGBoost \\
 & & \small $R^2 = 0.553$, \small MSE = 8.037 & \small $R^2 = 0.918$, \small MSE = 37.443 & \small $R^2 = 0.902$, \small MSE = 3.365 \\
\cmidrule(lr){2-5}
 & \multirow{2}{*}{Two-Shot} & XGBoost & Gradient Boosting & Gradient Boosting \\
 & & \small $R^2 = 0.866$, \small MSE = 6.322 & \small $R^2 = 0.950$, \small MSE = 18.687 & \small $R^2 = 0.801$, \small MSE = 2.950 \\
\bottomrule
\end{tabular}
\caption{Top Regression Models with $R^2$ and MSE for Each Language Model and Task}
\label{tab:top_regression_models_with_metrics}
\end{table*}

\section{Results}

\subsection{Regression Model Predictions}
This section explores factors influencing multilingual model performance by addressing three questions. First, we assess which regression models best predict performance, using R-squared ($R^2$) and Mean Squared Error (MSE) for F1 and sacreBLEU scores. Next, we identify key features driving model success. Finally, we examine how factors like geographical proximity, socio-linguistic context, and resource availability affect prediction accuracy, providing a comprehensive view of elements shaping model effectiveness.

Table \ref{tab:top_regression_models_with_metrics} presents the top-performing regression models for each language model and task setup, showing the best $R^2$ and Mean Squared Error (MSE) values. The detailed performance of various regression models can be found in Appendix~\ref{sec:appendix} (Tables \ref{tab:zero_shot_classification_results} and \ref{tab:two_shot_classification_results} for classification tasks, and Tables \ref{tab:zero_shot_generation_results} and \ref{tab:two_shot_generation_results} for generation tasks.)

Simpler models like SVR, K-Nearest Neighbors, and Lasso Regression generally performed poorly, often yielding negative $R^2$ scores and higher MSE values, indicating their limited ability to capture the complex interactions in the data. Linear models assume a straightforward proportional relationship between input features and the target variable, which was not effective here. In contrast, ensemble models such as Random Forest, Gradient Boosting, and XGBoost consistently excelled, demonstrating strong predictive performance across all tasks. These models achieved high $R^2$ scores and low MSE values, indicating that the \textit{relationships between features and performance metrics in multilingual language models are complex and non-linear with higher-order interactions}. 

Furthermore, the very low Mean Squared Error (MSE) values achieved by the best-performing regression models indicate that the features analyzed in this study are comprehensive and highly predictive of the model behavior. This low error rate suggests that \textit{there are no significant additional features with a high impact on model performance that were left out of the analysis.} The completeness of the set of features implies that we have effectively captured the key factors driving the performance of multilingual language models, thus providing a robust framework for understanding and predicting their behavior.

\subsection{Feature Importance Analysis}

To quantify the contribution of each feature to the performance of multilingual language models, we employed SHAP values, a powerful method for explaining individual predictions by measuring the marginal contribution of each feature, making it particularly suitable for complex models with non-linear interactions. In our analysis, SHAP values were used to rank the importance of various features, providing insights into which factors had the most significant impact on model performance across both classification and translation tasks. This method allowed us to understand the underlying drivers of performance disparities in multilingual models.

\begin{figure*}[!t]
    \centering
    \includegraphics[width=0.85\textwidth]{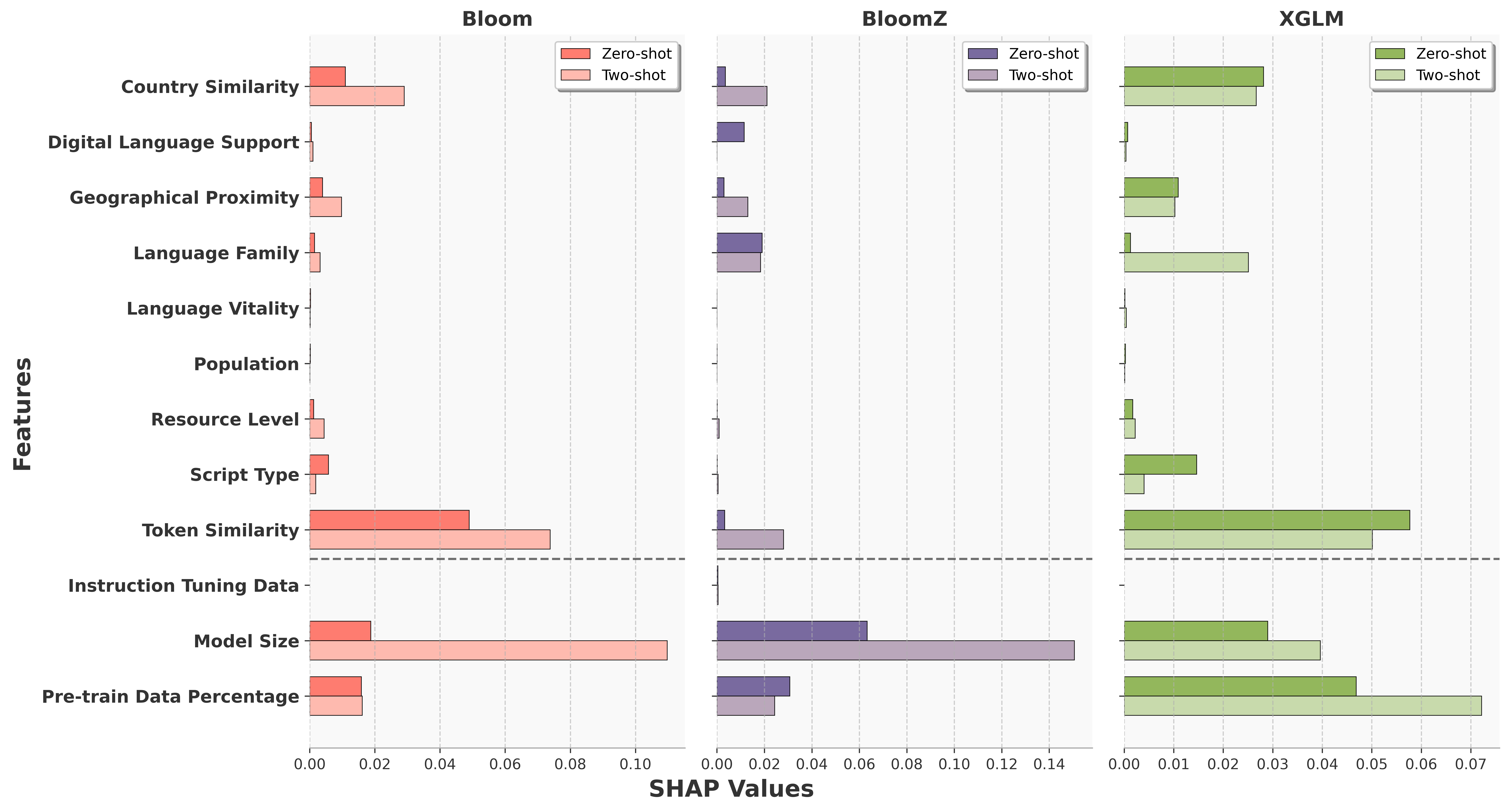}
    \caption{SHAP values for Zero-shot and Two-shot {\bf Classification} tasks across different models.}
    \label{fig:classification}
\end{figure*}

\begin{figure*}[!t]
    \centering
    \includegraphics[width=0.85\textwidth]{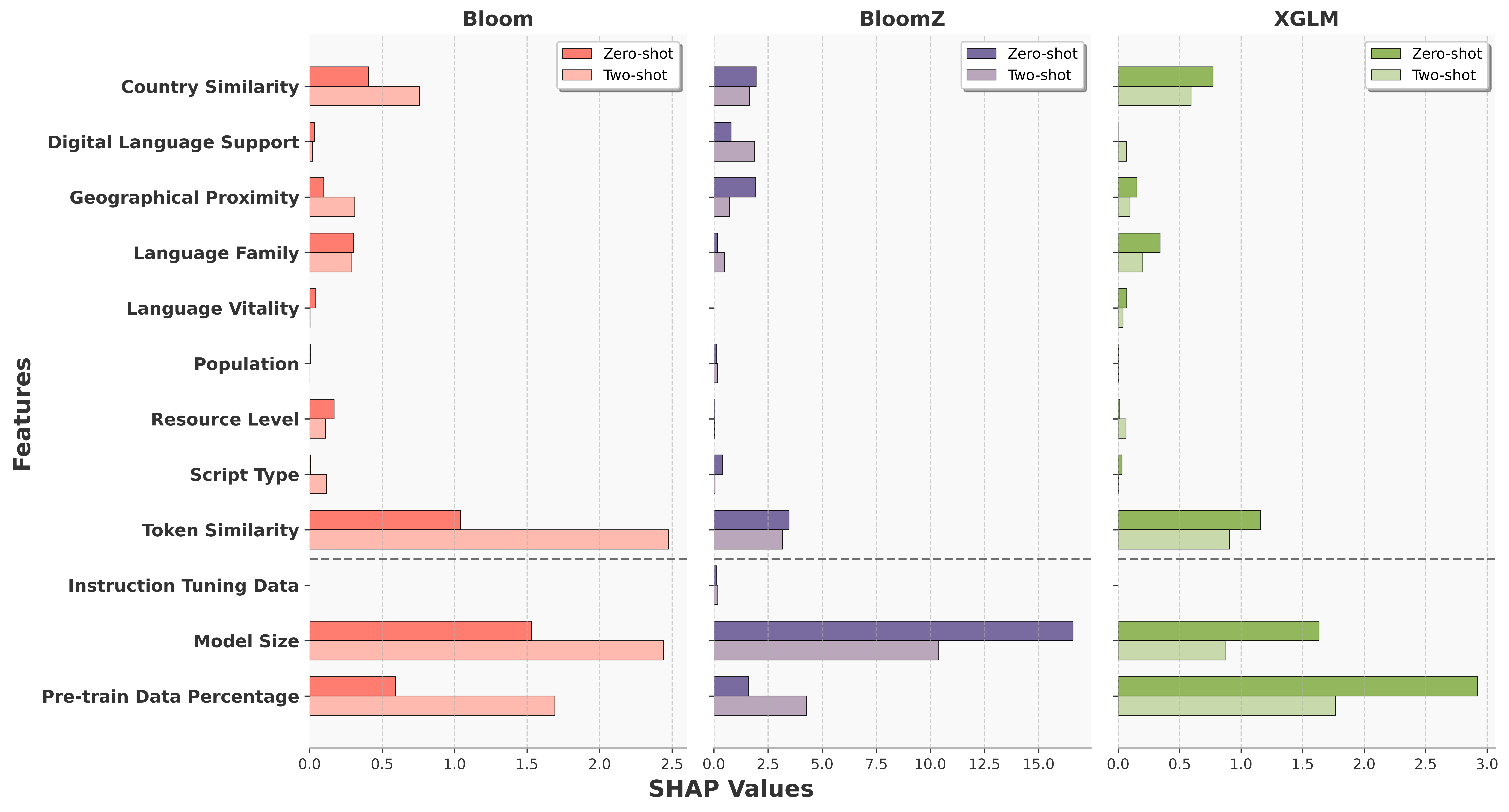}
    \caption{SHAP values for Zero-shot and Two-shot {\bf Generation} tasks across different models.}
    \label{fig:generation}
\end{figure*}

In both classification and generation tasks, as illustrated in Figures \ref{fig:classification} and \ref{fig:generation}, key features such as \textit{Token Similarity, Model Size, Pre-train Data Percentage, and Country Similarity consistently emerged as significant predictors of model performance across different settings.} Among these, Model Size was the most important feature in three out of six classification model setups and in three instances in generation tasks. Token Similarity was identified as a key feature twice in classification and once in generation, while Pre-train Data Percentage appeared as the most important feature once in classification and twice in generation. These findings suggest that focusing on these critical features can provide valuable insights into optimizing and improving the performance of multilingual language models.

\subsubsection{Model Features}
The model features—such as Pre-train Data Percentage, Instruction Tuning Data (specific to BloomZ), and Model Size—are crucial determinants of multilingual language model performance. 

Pre-train Data Percentage consistently emerged as a significant factor across both classification and generation tasks, as evidenced by its high SHAP values. This suggests that models are better equipped to capture linguistic nuances and achieve higher performance when more training data is available. The analysis highlights the importance of increasing pre-training data, particularly for underrepresented languages, to enhance the model’s ability to understand and generate language effectively.

Model Size also plays a critical role in determining performance. Larger models, with their increased number of parameters, have a greater capacity to learn complex patterns and relationships within the data, which is reflected in the consistently high SHAP values for this feature across various tasks. While larger models offer the advantage of more accurate predictions and higher-quality outputs, they also come with trade-offs, including higher computational demands and longer training times, which need to be considered when scaling up model sizes.

In contrast, Instruction Tuning Data—a feature unique to BloomZ—showed very low SHAP values, indicating its minimal impact on the model’s performance. This suggests that \textit{the model’s effectiveness is more strongly influenced by the amount of pre-training data rather than the fine-tuning process.} The analysis implies that while fine-tuning can refine a model’s capabilities, the scope and quality of pre-training data are far more critical in determining the overall effectiveness of the model, particularly in multilingual contexts.

\subsubsection{Geographical and Country Similarity}
The analysis of geographical proximity and country similarity revealed varying impacts on the performance of multilingual language models. While geographical proximity had a relatively modest influence, their SHAP values indicated that they still provided valuable context by capturing regional linguistic variations that could affect model predictions. For instance, languages spoken in geographically close regions might share linguistic characteristics that models can leverage for improved performance, even if these features were less important compared to others like Model Size and Token Similarity.

In contrast, country similarity had a more pronounced effect, frequently ranking among the top four features. The overlap of countries where languages are spoken often implies \textit{shared cultural and linguistic traits \cite{fishman1972language}, which multilingual models can utilize to enhance their predictions.} This suggests that languages with higher country similarity benefit from shared linguistic resources and transfer learning, thereby improving model performance.

The lower significance of geographical proximity might stem from the fact that geographical proximity does not always correlate with linguistic similarity. However, the stronger impact of country similarity, which directly relates to shared cultural and linguistic traits, underscores the importance of sociolinguistic factors in model performance. 

\subsubsection{Linguistic Features}
The impact of linguistic features, specifically Language Family and Script, on the performance of multilingual language models was analyzed, but the SHAP values indicated that these features had a relatively minor effect.

For Language Family, the SHAP values across both classification and generation tasks were generally low, suggesting that this feature did not significantly influence model performance. Although linguistic relatedness can facilitate transfer learning, the results imply that other features capture more crucial aspects of language modeling. Similarly, the Script feature also showed low importance according to the SHAP values. However, it is worth noting that Script type can indirectly influence model performance through its impact on Token Similarity.

\subsubsection{Token Similarity}
Token similarity emerged as one of the most crucial features influencing the performance of multilingual language models across both classification and generation tasks. This feature measures the overlap and similarity of tokens between different languages, providing a direct insight into how well the model can generalize and transfer learned knowledge from one language to another.

\textit{The consistent importance of token similarity across both tasks highlights its role in facilitating transfer learning and generalization in multilingual models.} Languages with high token similarity allow the model to reuse and adapt learned representations effectively, reducing the need for extensive language-specific training data. This finding emphasizes the value of incorporating token similarity measures when designing and evaluating multilingual language models.

Moreover, the high SHAP values associated with token similarity suggest that future improvements in multilingual models could focus on enhancing token representation and alignment across languages. Techniques such as multilingual token embeddings and shared subword tokenization strategies could further improve model performance by maximizing token overlap and similarity.

\subsubsection{Resource-Related Features}
Resource-related features, including Population, Language Vitality, Digital Language Support, and Resource Level, collectively capture the socio-linguistic context and the availability of digital resources for each language, factors which can influence model training and performance.

In our analysis, Population, referring to the number of speakers of a language, consistently showed very low SHAP values, indicating minimal impact on model performance. This suggests that while a larger speaker base might correlate with greater resource availability, it does not directly drive model success. Similarly, Language Vitality, which measures the robustness or endangerment of a language, also exhibited low SHAP values. This implies that even languages with lower vitality can achieve high model performance if they have sufficient high-quality training data.

Digital Language Support, which assesses the extent of digital resources available for a language, displayed moderate SHAP values in the BloomZ model but low values in others, indicating that its impact varies by model and is not a dominant factor overall. Resource Level, which reflects the availability of linguistic resources and data, also showed relatively low SHAP values.

Overall, \textit{while resource-related features can influence the availability of datasets for training language models, their direct impact on model performance is limited.} 

\section{Discussion}
The results of this study provide valuable insights into the factors that drive the performance of multilingual language models across classification and generation tasks. 

\paragraph{Ensemble Models and Feature Complexity:}
\begin{itemize}
    \item Ensemble models (Random Forest, Gradient Boosting, XGBoost) outperformed simpler linear models (SVR, Lasso Regression) across both classification and generation tasks.
    \item These models are better at capturing complex, non-linear interactions between features, highlighting the intricate relationships in multilingual language models.
\end{itemize}

\paragraph{Critical Role of Model Features:}
\begin{itemize}
    \item Pre-train Data Percentage and Model Size emerged as the most influential factors in model performance.
    \item Larger models showed superior performance due to their ability to learn complex data patterns.
    \item Instruction Tuning Data had minimal impact on performance, indicating that pre-training data is more crucial than fine-tuning.
\end{itemize}

\paragraph{Importance of Token Similarity:}
\begin{itemize}
    \item Token similarity was a top predictor of model performance, facilitating effective transfer learning and generalization.
    \item Optimizing token representation and alignment across languages could further improve multilingual model performance.
\end{itemize}

\paragraph{Geographical and Sociolinguistic Context:}
\begin{itemize}
    \item While geographical proximity had a modest impact, country similarity was more significant in driving model performance.
    \item Shared cultural and linguistic traits across countries enhance model predictions, emphasizing the importance of considering sociolinguistic factors.
\end{itemize}

\paragraph{Resource-Related Features:}
\begin{itemize}
    \item Features like Population, Language Vitality, Digital Language Support, and Resource Level had limited direct impact on model performance.
    \item Although, the availability of resources is essential for providing high-quality training data, they are not primary determinants of model success.
\end{itemize}

\section{Conclusion}
This study offers a detailed analysis of the factors influencing multilingual language model performance across classification and generation tasks. Our findings show that performance is shaped by complex, non-linear interactions among features. Key factors include pre-train data percentage and model size, which significantly affect effectiveness. Token similarity enhances cross-lingual transfer learning, while country similarity highlights the role of shared cultural and linguistic contexts. Resource-related features like population and digital support showed limited direct impact but remain useful for understanding data availability and training strategies. These insights are crucial for developing more equitable multilingual models, especially for underrepresented languages.

\section{Limitation}
This study, while comprehensive, has several limitations. The analysis is focused on specific models (Bloom, BloomZ, and XGLM), which may limit generalizability to other architectures. Additionally, reliance on SHAP values might overlook complex interactions between features. The datasets (SIB-200 and Flores-200) cover many languages but may not fully capture dialectal diversity, and computational constraints restricted testing to a range of model sizes. Future work could address these aspects by exploring more models, diverse datasets, and further feature interactions.

\section*{Acknowledgments}
This work was supported by the National Artificial Intelligence Research Resource (NAIRR) Pilot, funded by the National Science Foundation under award No. 240158.

\bibliography{custom}

\begin{thebibliography}{40}
\providecommand{\natexlab}[1]{#1}

\bibitem[{Adelani et~al.(2024)Adelani, Liu, Shen, Vassilyev, Alabi, Mao, Gao, and Lee}]{adelani-etal-2024-sib}
David Adelani, Hannah Liu, Xiaoyu Shen, Nikita Vassilyev, Jesujoba Alabi, Yanke Mao, Haonan Gao, and En-Shiun Lee. 2024.
\newblock \href {https://aclanthology.org/2024.eacl-long.14} {{SIB}-200: A simple, inclusive, and big evaluation dataset for topic classification in 200+ languages and dialects}.
\newblock In \emph{Proceedings of the 18th Conference of the European Chapter of the Association for Computational Linguistics (Volume 1: Long Papers)}, pages 226--245, St. Julian{'}s, Malta. Association for Computational Linguistics.

\bibitem[{Bagheri~Nezhad and Agrawal(2024)}]{bagheri-nezhad-agrawal-2024-drives}
Sina Bagheri~Nezhad and Ameeta Agrawal. 2024.
\newblock \href {https://doi.org/10.18653/v1/2024.vardial-1.2} {What drives performance in multilingual language models?}
\newblock In \emph{Proceedings of the Eleventh Workshop on NLP for Similar Languages, Varieties, and Dialects (VarDial 2024)}, pages 16--27, Mexico City, Mexico. Association for Computational Linguistics.

\bibitem[{BigScience et~al.(2023)BigScience, :, Scao, Fan, Akiki, Pavlick, Ilić, Hesslow, Castagné, Luccioni, Yvon, Gallé, Tow, Rush, Biderman, Webson, Ammanamanchi, Wang, Sagot, Muennighoff, del Moral, Ruwase, Bawden, Bekman, McMillan-Major, Beltagy, Nguyen, Saulnier, Tan, Suarez, Sanh, Laurençon, Jernite, Launay, Mitchell, Raffel, Gokaslan, Simhi, Soroa, Aji, Alfassy, Rogers, Nitzav, Xu, Mou, Emezue, Klamm, Leong, van Strien, Adelani, Radev, Ponferrada, Levkovizh, Kim, Natan, Toni, Dupont, Kruszewski, Pistilli, Elsahar, Benyamina, Tran, Yu, Abdulmumin, Johnson, Gonzalez-Dios, de~la Rosa, Chim, Dodge, Zhu, Chang, Frohberg, Tobing, Bhattacharjee, Almubarak, Chen, Lo, Werra, Weber, Phan, allal, Tanguy, Dey, Muñoz, Masoud, Grandury, Šaško, Huang, Coavoux, Singh, Jiang, Vu, Jauhar, Ghaleb, Subramani, Kassner, Khamis, Nguyen, Espejel, de~Gibert, Villegas, Henderson, Colombo, Amuok, Lhoest, Harliman, Bommasani, López, Ribeiro, Osei, Pyysalo, Nagel, Bose, Muhammad, Sharma, Longpre, Nikpoor, Silberberg, Pai,
  Zink, Torrent, Schick, Thrush, Danchev, Nikoulina, Laippala, Lepercq, Prabhu, Alyafeai, Talat, Raja, Heinzerling, Si, Taşar, Salesky, Mielke, Lee, Sharma, Santilli, Chaffin, Stiegler, Datta, Szczechla, Chhablani, Wang, Pandey, Strobelt, Fries, Rozen, Gao, Sutawika, Bari, Al-shaibani, Manica, Nayak, Teehan, Albanie, Shen, Ben-David, Bach, Kim, Bers, Fevry, Neeraj, Thakker, Raunak, Tang, Yong, Sun, Brody, Uri, Tojarieh, Roberts, Chung, Tae, Phang, Press, Li, Narayanan, Bourfoune, Casper, Rasley, Ryabinin, Mishra, Zhang, Shoeybi, Peyrounette, Patry, Tazi, Sanseviero, von Platen, Cornette, Lavallée, Lacroix, Rajbhandari, Gandhi, Smith, Requena, Patil, Dettmers, Baruwa, Singh, Cheveleva, Ligozat, Subramonian, Névéol, Lovering, Garrette, Tunuguntla, Reiter, Taktasheva, Voloshina, Bogdanov, Winata, Schoelkopf, Kalo, Novikova, Forde, Clive, Kasai, Kawamura, Hazan, Carpuat, Clinciu, Kim, Cheng, Serikov, Antverg, van~der Wal, Zhang, Zhang, Gehrmann, Mirkin, Pais, Shavrina, Scialom, Yun, Limisiewicz, Rieser,
  Protasov, Mikhailov, Pruksachatkun, Belinkov, Bamberger, Kasner, Rueda, Pestana, Feizpour, Khan, Faranak, Santos, Hevia, Unldreaj, Aghagol, Abdollahi, Tammour, HajiHosseini, Behroozi, Ajibade, Saxena, Ferrandis, McDuff, Contractor, Lansky, David, Kiela, Nguyen, Tan, Baylor, Ozoani, Mirza, Ononiwu, Rezanejad, Jones, Bhattacharya, Solaiman, Sedenko, Nejadgholi, Passmore, Seltzer, Sanz, Dutra, Samagaio, Elbadri, Mieskes, Gerchick, Akinlolu, McKenna, Qiu, Ghauri, Burynok, Abrar, Rajani, Elkott, Fahmy, Samuel, An, Kromann, Hao, Alizadeh, Shubber, Wang, Roy, Viguier, Le, Oyebade, Le, Yang, Nguyen, Kashyap, Palasciano, Callahan, Shukla, Miranda-Escalada, Singh, Beilharz, Wang, Brito, Zhou, Jain, Xu, Fourrier, Periñán, Molano, Yu, Manjavacas, Barth, Fuhrimann, Altay, Bayrak, Burns, Vrabec, Bello, Dash, Kang, Giorgi, Golde, Posada, Sivaraman, Bulchandani, Liu, Shinzato, de~Bykhovetz, Takeuchi, Pàmies, Castillo, Nezhurina, Sänger, Samwald, Cullan, Weinberg, Wolf, Mihaljcic, Liu, Freidank, Kang, Seelam, Dahlberg,
  Broad, Muellner, Fung, Haller, Chandrasekhar, Eisenberg, Martin, Canalli, Su, Su, Cahyawijaya, Garda, Deshmukh, Mishra, Kiblawi, Ott, Sang-aroonsiri, Kumar, Schweter, Bharati, Laud, Gigant, Kainuma, Kusa, Labrak, Bajaj, Venkatraman, Xu, Xu, Xu, Tan, Xie, Ye, Bras, Belkada, and Wolf}]{workshop2023Bloom176bparameteropenaccessmultilingual}
BigScience, :, Teven~Le Scao, Angela Fan, Christopher Akiki, Ellie Pavlick, Suzana Ilić, Daniel Hesslow, Roman Castagné, Alexandra~Sasha Luccioni, François Yvon, Matthias Gallé, Jonathan Tow, Alexander~M. Rush, Stella Biderman, Albert Webson, Pawan~Sasanka Ammanamanchi, Thomas Wang, Benoît Sagot, Niklas Muennighoff, Albert~Villanova del Moral, Olatunji Ruwase, Rachel Bawden, Stas Bekman, Angelina McMillan-Major, Iz~Beltagy, Huu Nguyen, Lucile Saulnier, Samson Tan, Pedro~Ortiz Suarez, Victor Sanh, Hugo Laurençon, Yacine Jernite, Julien Launay, Margaret Mitchell, Colin Raffel, Aaron Gokaslan, Adi Simhi, Aitor Soroa, Alham~Fikri Aji, Amit Alfassy, Anna Rogers, Ariel~Kreisberg Nitzav, Canwen Xu, Chenghao Mou, Chris Emezue, Christopher Klamm, Colin Leong, Daniel van Strien, David~Ifeoluwa Adelani, Dragomir Radev, Eduardo~González Ponferrada, Efrat Levkovizh, Ethan Kim, Eyal~Bar Natan, Francesco~De Toni, Gérard Dupont, Germán Kruszewski, Giada Pistilli, Hady Elsahar, Hamza Benyamina, Hieu Tran, Ian Yu,
  Idris Abdulmumin, Isaac Johnson, Itziar Gonzalez-Dios, Javier de~la Rosa, Jenny Chim, Jesse Dodge, Jian Zhu, Jonathan Chang, Jörg Frohberg, Joseph Tobing, Joydeep Bhattacharjee, Khalid Almubarak, Kimbo Chen, Kyle Lo, Leandro~Von Werra, Leon Weber, Long Phan, Loubna~Ben allal, Ludovic Tanguy, Manan Dey, Manuel~Romero Muñoz, Maraim Masoud, María Grandury, Mario Šaško, Max Huang, Maximin Coavoux, Mayank Singh, Mike Tian-Jian Jiang, Minh~Chien Vu, Mohammad~A. Jauhar, Mustafa Ghaleb, Nishant Subramani, Nora Kassner, Nurulaqilla Khamis, Olivier Nguyen, Omar Espejel, Ona de~Gibert, Paulo Villegas, Peter Henderson, Pierre Colombo, Priscilla Amuok, Quentin Lhoest, Rheza Harliman, Rishi Bommasani, Roberto~Luis López, Rui Ribeiro, Salomey Osei, Sampo Pyysalo, Sebastian Nagel, Shamik Bose, Shamsuddeen~Hassan Muhammad, Shanya Sharma, Shayne Longpre, Somaieh Nikpoor, Stanislav Silberberg, Suhas Pai, Sydney Zink, Tiago~Timponi Torrent, Timo Schick, Tristan Thrush, Valentin Danchev, Vassilina Nikoulina, Veronika
  Laippala, Violette Lepercq, Vrinda Prabhu, Zaid Alyafeai, Zeerak Talat, Arun Raja, Benjamin Heinzerling, Chenglei Si, Davut~Emre Taşar, Elizabeth Salesky, Sabrina~J. Mielke, Wilson~Y. Lee, Abheesht Sharma, Andrea Santilli, Antoine Chaffin, Arnaud Stiegler, Debajyoti Datta, Eliza Szczechla, Gunjan Chhablani, Han Wang, Harshit Pandey, Hendrik Strobelt, Jason~Alan Fries, Jos Rozen, Leo Gao, Lintang Sutawika, M~Saiful Bari, Maged~S. Al-shaibani, Matteo Manica, Nihal Nayak, Ryan Teehan, Samuel Albanie, Sheng Shen, Srulik Ben-David, Stephen~H. Bach, Taewoon Kim, Tali Bers, Thibault Fevry, Trishala Neeraj, Urmish Thakker, Vikas Raunak, Xiangru Tang, Zheng-Xin Yong, Zhiqing Sun, Shaked Brody, Yallow Uri, Hadar Tojarieh, Adam Roberts, Hyung~Won Chung, Jaesung Tae, Jason Phang, Ofir Press, Conglong Li, Deepak Narayanan, Hatim Bourfoune, Jared Casper, Jeff Rasley, Max Ryabinin, Mayank Mishra, Minjia Zhang, Mohammad Shoeybi, Myriam Peyrounette, Nicolas Patry, Nouamane Tazi, Omar Sanseviero, Patrick von Platen, Pierre
  Cornette, Pierre~François Lavallée, Rémi Lacroix, Samyam Rajbhandari, Sanchit Gandhi, Shaden Smith, Stéphane Requena, Suraj Patil, Tim Dettmers, Ahmed Baruwa, Amanpreet Singh, Anastasia Cheveleva, Anne-Laure Ligozat, Arjun Subramonian, Aurélie Névéol, Charles Lovering, Dan Garrette, Deepak Tunuguntla, Ehud Reiter, Ekaterina Taktasheva, Ekaterina Voloshina, Eli Bogdanov, Genta~Indra Winata, Hailey Schoelkopf, Jan-Christoph Kalo, Jekaterina Novikova, Jessica~Zosa Forde, Jordan Clive, Jungo Kasai, Ken Kawamura, Liam Hazan, Marine Carpuat, Miruna Clinciu, Najoung Kim, Newton Cheng, Oleg Serikov, Omer Antverg, Oskar van~der Wal, Rui Zhang, Ruochen Zhang, Sebastian Gehrmann, Shachar Mirkin, Shani Pais, Tatiana Shavrina, Thomas Scialom, Tian Yun, Tomasz Limisiewicz, Verena Rieser, Vitaly Protasov, Vladislav Mikhailov, Yada Pruksachatkun, Yonatan Belinkov, Zachary Bamberger, Zdeněk Kasner, Alice Rueda, Amanda Pestana, Amir Feizpour, Ammar Khan, Amy Faranak, Ana Santos, Anthony Hevia, Antigona Unldreaj,
  Arash Aghagol, Arezoo Abdollahi, Aycha Tammour, Azadeh HajiHosseini, Bahareh Behroozi, Benjamin Ajibade, Bharat Saxena, Carlos~Muñoz Ferrandis, Daniel McDuff, Danish Contractor, David Lansky, Davis David, Douwe Kiela, Duong~A. Nguyen, Edward Tan, Emi Baylor, Ezinwanne Ozoani, Fatima Mirza, Frankline Ononiwu, Habib Rezanejad, Hessie Jones, Indrani Bhattacharya, Irene Solaiman, Irina Sedenko, Isar Nejadgholi, Jesse Passmore, Josh Seltzer, Julio~Bonis Sanz, Livia Dutra, Mairon Samagaio, Maraim Elbadri, Margot Mieskes, Marissa Gerchick, Martha Akinlolu, Michael McKenna, Mike Qiu, Muhammed Ghauri, Mykola Burynok, Nafis Abrar, Nazneen Rajani, Nour Elkott, Nour Fahmy, Olanrewaju Samuel, Ran An, Rasmus Kromann, Ryan Hao, Samira Alizadeh, Sarmad Shubber, Silas Wang, Sourav Roy, Sylvain Viguier, Thanh Le, Tobi Oyebade, Trieu Le, Yoyo Yang, Zach Nguyen, Abhinav~Ramesh Kashyap, Alfredo Palasciano, Alison Callahan, Anima Shukla, Antonio Miranda-Escalada, Ayush Singh, Benjamin Beilharz, Bo~Wang, Caio Brito, Chenxi Zhou,
  Chirag Jain, Chuxin Xu, Clémentine Fourrier, Daniel~León Periñán, Daniel Molano, Dian Yu, Enrique Manjavacas, Fabio Barth, Florian Fuhrimann, Gabriel Altay, Giyaseddin Bayrak, Gully Burns, Helena~U. Vrabec, Imane Bello, Ishani Dash, Jihyun Kang, John Giorgi, Jonas Golde, Jose~David Posada, Karthik~Rangasai Sivaraman, Lokesh Bulchandani, Lu~Liu, Luisa Shinzato, Madeleine~Hahn de~Bykhovetz, Maiko Takeuchi, Marc Pàmies, Maria~A Castillo, Marianna Nezhurina, Mario Sänger, Matthias Samwald, Michael Cullan, Michael Weinberg, Michiel~De Wolf, Mina Mihaljcic, Minna Liu, Moritz Freidank, Myungsun Kang, Natasha Seelam, Nathan Dahlberg, Nicholas~Michio Broad, Nikolaus Muellner, Pascale Fung, Patrick Haller, Ramya Chandrasekhar, Renata Eisenberg, Robert Martin, Rodrigo Canalli, Rosaline Su, Ruisi Su, Samuel Cahyawijaya, Samuele Garda, Shlok~S Deshmukh, Shubhanshu Mishra, Sid Kiblawi, Simon Ott, Sinee Sang-aroonsiri, Srishti Kumar, Stefan Schweter, Sushil Bharati, Tanmay Laud, Théo Gigant, Tomoya Kainuma,
  Wojciech Kusa, Yanis Labrak, Yash~Shailesh Bajaj, Yash Venkatraman, Yifan Xu, Yingxin Xu, Yu~Xu, Zhe Tan, Zhongli Xie, Zifan Ye, Mathilde Bras, Younes Belkada, and Thomas Wolf. 2023.
\newblock \href {https://arxiv.org/abs/2211.05100} {Bloom: A 176b-parameter open-access multilingual language model}.
\newblock \emph{Preprint}, arXiv:2211.05100.

\bibitem[{Breiman(2001)}]{breiman2001random}
Leo Breiman. 2001.
\newblock Random forests.
\newblock \emph{Machine learning}, 45(1):5--32.

\bibitem[{Brown et~al.(2020)Brown, Mann, Ryder, Subbiah, Kaplan, Dhariwal, Neelakantan, Shyam, Sastry, Askell, Agarwal, Herbert-Voss, Krueger, Henighan, Child, Ramesh, Ziegler, Wu, Winter, Hesse, Chen, Sigler, Litwin, Gray, Chess, Clark, Berner, McCandlish, Radford, Sutskever, and Amodei}]{NEURIPS2020_1457c0d6}
Tom Brown, Benjamin Mann, Nick Ryder, Melanie Subbiah, Jared~D Kaplan, Prafulla Dhariwal, Arvind Neelakantan, Pranav Shyam, Girish Sastry, Amanda Askell, Sandhini Agarwal, Ariel Herbert-Voss, Gretchen Krueger, Tom Henighan, Rewon Child, Aditya Ramesh, Daniel Ziegler, Jeffrey Wu, Clemens Winter, Chris Hesse, Mark Chen, Eric Sigler, Mateusz Litwin, Scott Gray, Benjamin Chess, Jack Clark, Christopher Berner, Sam McCandlish, Alec Radford, Ilya Sutskever, and Dario Amodei. 2020.
\newblock \href {https://proceedings.neurips.cc/paper_files/paper/2020/file/1457c0d6bfcb4967418bfb8ac142f64a-Paper.pdf} {Language models are few-shot learners}.
\newblock In \emph{Advances in Neural Information Processing Systems}, volume~33, pages 1877--1901. Curran Associates, Inc.

\bibitem[{Chau and Smith(2021)}]{chau-smith-2021-specializing}
Ethan~C. Chau and Noah~A. Smith. 2021.
\newblock \href {https://doi.org/10.18653/v1/2021.mrl-1.5} {Specializing multilingual language models: An empirical study}.
\newblock In \emph{Proceedings of the 1st Workshop on Multilingual Representation Learning}, pages 51--61, Punta Cana, Dominican Republic. Association for Computational Linguistics.

\bibitem[{Chen and Guestrin(2016)}]{chen2016xgboost}
Tianqi Chen and Carlos Guestrin. 2016.
\newblock Xgboost: A scalable tree boosting system.
\newblock In \emph{Proceedings of the 22nd ACM SIGKDD International Conference on Knowledge Discovery and Data Mining}, pages 785--794.

\bibitem[{Conneau et~al.(2020)Conneau, Khandelwal, Goyal, Chaudhary, Wenzek, Guzm{\'a}n, Grave, Ott, Zettlemoyer, and Stoyanov}]{conneau-etal-2020-unsupervised}
Alexis Conneau, Kartikay Khandelwal, Naman Goyal, Vishrav Chaudhary, Guillaume Wenzek, Francisco Guzm{\'a}n, Edouard Grave, Myle Ott, Luke Zettlemoyer, and Veselin Stoyanov. 2020.
\newblock \href {https://doi.org/10.18653/v1/2020.acl-main.747} {Unsupervised cross-lingual representation learning at scale}.
\newblock In \emph{Proceedings of the 58th Annual Meeting of the Association for Computational Linguistics}, pages 8440--8451, Online. Association for Computational Linguistics.

\bibitem[{Devlin et~al.(2019)Devlin, Chang, Lee, and Toutanova}]{devlin-etal-2019-bert}
Jacob Devlin, Ming-Wei Chang, Kenton Lee, and Kristina Toutanova. 2019.
\newblock \href {https://doi.org/10.18653/v1/N19-1423} {{BERT}: Pre-training of deep bidirectional transformers for language understanding}.
\newblock In \emph{Proceedings of the 2019 Conference of the North {A}merican Chapter of the Association for Computational Linguistics: Human Language Technologies, Volume 1 (Long and Short Papers)}, pages 4171--4186, Minneapolis, Minnesota. Association for Computational Linguistics.

\bibitem[{Dubey et~al.(2024)Dubey, Jauhri, Pandey, Kadian, Al-Dahle, Letman, Mathur, Schelten, Yang, Fan, Goyal, Hartshorn, Yang, Mitra, Sravankumar, Korenev, Hinsvark, Rao, Zhang, Rodriguez, Gregerson, Spataru, Roziere, Biron, Tang, Chern, Caucheteux, Nayak, Bi, Marra, McConnell, Keller, Touret, Wu, Wong, Ferrer, Nikolaidis, Allonsius, Song, Pintz, Livshits, Esiobu, Choudhary, Mahajan, Garcia-Olano, Perino, Hupkes, Lakomkin, AlBadawy, Lobanova, Dinan, Smith, Radenovic, Zhang, Synnaeve, Lee, Anderson, Nail, Mialon, Pang, Cucurell, Nguyen, Korevaar, Xu, Touvron, Zarov, Ibarra, Kloumann, Misra, Evtimov, Copet, Lee, Geffert, Vranes, Park, Mahadeokar, Shah, van~der Linde, Billock, Hong, Lee, Fu, Chi, Huang, Liu, Wang, Yu, Bitton, Spisak, Park, Rocca, Johnstun, Saxe, Jia, Alwala, Upasani, Plawiak, Li, Heafield, Stone, El-Arini, Iyer, Malik, Chiu, Bhalla, Rantala-Yeary, van~der Maaten, Chen, Tan, Jenkins, Martin, Madaan, Malo, Blecher, Landzaat, de~Oliveira, Muzzi, Pasupuleti, Singh, Paluri, Kardas, Oldham, Rita,
  Pavlova, Kambadur, Lewis, Si, Singh, Hassan, Goyal, Torabi, Bashlykov, Bogoychev, Chatterji, Duchenne, Çelebi, Alrassy, Zhang, Li, Vasic, Weng, Bhargava, Dubal, Krishnan, Koura, Xu, He, Dong, Srinivasan, Ganapathy, Calderer, Cabral, Stojnic, Raileanu, Girdhar, Patel, Sauvestre, Polidoro, Sumbaly, Taylor, Silva, Hou, Wang, Hosseini, Chennabasappa, Singh, Bell, Kim, Edunov, Nie, Narang, Raparthy, Shen, Wan, Bhosale, Zhang, Vandenhende, Batra, Whitman, Sootla, Collot, Gururangan, Borodinsky, Herman, Fowler, Sheasha, Georgiou, Scialom, Speckbacher, Mihaylov, Xiao, Karn, Goswami, Gupta, Ramanathan, Kerkez, Gonguet, Do, Vogeti, Petrovic, Chu, Xiong, Fu, Meers, Martinet, Wang, Tan, Xie, Jia, Wang, Goldschlag, Gaur, Babaei, Wen, Song, Zhang, Li, Mao, Coudert, Yan, Chen, Papakipos, Singh, Grattafiori, Jain, Kelsey, Shajnfeld, Gangidi, Victoria, Goldstand, Menon, Sharma, Boesenberg, Vaughan, Baevski, Feinstein, Kallet, Sangani, Yunus, Lupu, Alvarado, Caples, Gu, Ho, Poulton, Ryan, Ramchandani, Franco, Saraf,
  Chowdhury, Gabriel, Bharambe, Eisenman, Yazdan, James, Maurer, Leonhardi, Huang, Loyd, Paola, Paranjape, Liu, Wu, Ni, Hancock, Wasti, Spence, Stojkovic, Gamido, Montalvo, Parker, Burton, Mejia, Wang, Kim, Zhou, Hu, Chu, Cai, Tindal, Feichtenhofer, Civin, Beaty, Kreymer, Li, Wyatt, Adkins, Xu, Testuggine, David, Parikh, Liskovich, Foss, Wang, Le, Holland, Dowling, Jamil, Montgomery, Presani, Hahn, Wood, Brinkman, Arcaute, Dunbar, Smothers, Sun, Kreuk, Tian, Ozgenel, Caggioni, Guzmán, Kanayet, Seide, Florez, Schwarz, Badeer, Swee, Halpern, Thattai, Herman, Sizov, Guangyi, Zhang, Lakshminarayanan, Shojanazeri, Zou, Wang, Zha, Habeeb, Rudolph, Suk, Aspegren, Goldman, Molybog, Tufanov, Veliche, Gat, Weissman, Geboski, Kohli, Asher, Gaya, Marcus, Tang, Chan, Zhen, Reizenstein, Teboul, Zhong, Jin, Yang, Cummings, Carvill, Shepard, McPhie, Torres, Ginsburg, Wang, Wu, U, Saxena, Prasad, Khandelwal, Zand, Matosich, Veeraraghavan, Michelena, Li, Huang, Chawla, Lakhotia, Huang, Chen, Garg, A, Silva, Bell, Zhang, Guo,
  Yu, Moshkovich, Wehrstedt, Khabsa, Avalani, Bhatt, Tsimpoukelli, Mankus, Hasson, Lennie, Reso, Groshev, Naumov, Lathi, Keneally, Seltzer, Valko, Restrepo, Patel, Vyatskov, Samvelyan, Clark, Macey, Wang, Hermoso, Metanat, Rastegari, Bansal, Santhanam, Parks, White, Bawa, Singhal, Egebo, Usunier, Laptev, Dong, Zhang, Cheng, Chernoguz, Hart, Salpekar, Kalinli, Kent, Parekh, Saab, Balaji, Rittner, Bontrager, Roux, Dollar, Zvyagina, Ratanchandani, Yuvraj, Liang, Alao, Rodriguez, Ayub, Murthy, Nayani, Mitra, Li, Hogan, Battey, Wang, Maheswari, Howes, Rinott, Bondu, Datta, Chugh, Hunt, Dhillon, Sidorov, Pan, Verma, Yamamoto, Ramaswamy, Lindsay, Lindsay, Feng, Lin, Zha, Shankar, Zhang, Zhang, Wang, Agarwal, Sajuyigbe, Chintala, Max, Chen, Kehoe, Satterfield, Govindaprasad, Gupta, Cho, Virk, Subramanian, Choudhury, Goldman, Remez, Glaser, Best, Kohler, Robinson, Li, Zhang, Matthews, Chou, Shaked, Vontimitta, Ajayi, Montanez, Mohan, Kumar, Mangla, Ionescu, Poenaru, Mihailescu, Ivanov, Li, Wang, Jiang, Bouaziz,
  Constable, Tang, Wang, Wu, Wang, Xia, Wu, Gao, Chen, Hu, Jia, Qi, Li, Zhang, Zhang, Adi, Nam, Yu, Wang, Hao, Qian, He, Rait, DeVito, Rosnbrick, Wen, Yang, and Zhao}]{dubey2024llama3herdmodels}
Abhimanyu Dubey, Abhinav Jauhri, Abhinav Pandey, Abhishek Kadian, Ahmad Al-Dahle, Aiesha Letman, Akhil Mathur, Alan Schelten, Amy Yang, Angela Fan, Anirudh Goyal, Anthony Hartshorn, Aobo Yang, Archi Mitra, Archie Sravankumar, Artem Korenev, Arthur Hinsvark, Arun Rao, Aston Zhang, Aurelien Rodriguez, Austen Gregerson, Ava Spataru, Baptiste Roziere, Bethany Biron, Binh Tang, Bobbie Chern, Charlotte Caucheteux, Chaya Nayak, Chloe Bi, Chris Marra, Chris McConnell, Christian Keller, Christophe Touret, Chunyang Wu, Corinne Wong, Cristian~Canton Ferrer, Cyrus Nikolaidis, Damien Allonsius, Daniel Song, Danielle Pintz, Danny Livshits, David Esiobu, Dhruv Choudhary, Dhruv Mahajan, Diego Garcia-Olano, Diego Perino, Dieuwke Hupkes, Egor Lakomkin, Ehab AlBadawy, Elina Lobanova, Emily Dinan, Eric~Michael Smith, Filip Radenovic, Frank Zhang, Gabriel Synnaeve, Gabrielle Lee, Georgia~Lewis Anderson, Graeme Nail, Gregoire Mialon, Guan Pang, Guillem Cucurell, Hailey Nguyen, Hannah Korevaar, Hu~Xu, Hugo Touvron, Iliyan Zarov,
  Imanol~Arrieta Ibarra, Isabel Kloumann, Ishan Misra, Ivan Evtimov, Jade Copet, Jaewon Lee, Jan Geffert, Jana Vranes, Jason Park, Jay Mahadeokar, Jeet Shah, Jelmer van~der Linde, Jennifer Billock, Jenny Hong, Jenya Lee, Jeremy Fu, Jianfeng Chi, Jianyu Huang, Jiawen Liu, Jie Wang, Jiecao Yu, Joanna Bitton, Joe Spisak, Jongsoo Park, Joseph Rocca, Joshua Johnstun, Joshua Saxe, Junteng Jia, Kalyan~Vasuden Alwala, Kartikeya Upasani, Kate Plawiak, Ke~Li, Kenneth Heafield, Kevin Stone, Khalid El-Arini, Krithika Iyer, Kshitiz Malik, Kuenley Chiu, Kunal Bhalla, Lauren Rantala-Yeary, Laurens van~der Maaten, Lawrence Chen, Liang Tan, Liz Jenkins, Louis Martin, Lovish Madaan, Lubo Malo, Lukas Blecher, Lukas Landzaat, Luke de~Oliveira, Madeline Muzzi, Mahesh Pasupuleti, Mannat Singh, Manohar Paluri, Marcin Kardas, Mathew Oldham, Mathieu Rita, Maya Pavlova, Melanie Kambadur, Mike Lewis, Min Si, Mitesh~Kumar Singh, Mona Hassan, Naman Goyal, Narjes Torabi, Nikolay Bashlykov, Nikolay Bogoychev, Niladri Chatterji, Olivier
  Duchenne, Onur Çelebi, Patrick Alrassy, Pengchuan Zhang, Pengwei Li, Petar Vasic, Peter Weng, Prajjwal Bhargava, Pratik Dubal, Praveen Krishnan, Punit~Singh Koura, Puxin Xu, Qing He, Qingxiao Dong, Ragavan Srinivasan, Raj Ganapathy, Ramon Calderer, Ricardo~Silveira Cabral, Robert Stojnic, Roberta Raileanu, Rohit Girdhar, Rohit Patel, Romain Sauvestre, Ronnie Polidoro, Roshan Sumbaly, Ross Taylor, Ruan Silva, Rui Hou, Rui Wang, Saghar Hosseini, Sahana Chennabasappa, Sanjay Singh, Sean Bell, Seohyun~Sonia Kim, Sergey Edunov, Shaoliang Nie, Sharan Narang, Sharath Raparthy, Sheng Shen, Shengye Wan, Shruti Bhosale, Shun Zhang, Simon Vandenhende, Soumya Batra, Spencer Whitman, Sten Sootla, Stephane Collot, Suchin Gururangan, Sydney Borodinsky, Tamar Herman, Tara Fowler, Tarek Sheasha, Thomas Georgiou, Thomas Scialom, Tobias Speckbacher, Todor Mihaylov, Tong Xiao, Ujjwal Karn, Vedanuj Goswami, Vibhor Gupta, Vignesh Ramanathan, Viktor Kerkez, Vincent Gonguet, Virginie Do, Vish Vogeti, Vladan Petrovic, Weiwei Chu,
  Wenhan Xiong, Wenyin Fu, Whitney Meers, Xavier Martinet, Xiaodong Wang, Xiaoqing~Ellen Tan, Xinfeng Xie, Xuchao Jia, Xuewei Wang, Yaelle Goldschlag, Yashesh Gaur, Yasmine Babaei, Yi~Wen, Yiwen Song, Yuchen Zhang, Yue Li, Yuning Mao, Zacharie~Delpierre Coudert, Zheng Yan, Zhengxing Chen, Zoe Papakipos, Aaditya Singh, Aaron Grattafiori, Abha Jain, Adam Kelsey, Adam Shajnfeld, Adithya Gangidi, Adolfo Victoria, Ahuva Goldstand, Ajay Menon, Ajay Sharma, Alex Boesenberg, Alex Vaughan, Alexei Baevski, Allie Feinstein, Amanda Kallet, Amit Sangani, Anam Yunus, Andrei Lupu, Andres Alvarado, Andrew Caples, Andrew Gu, Andrew Ho, Andrew Poulton, Andrew Ryan, Ankit Ramchandani, Annie Franco, Aparajita Saraf, Arkabandhu Chowdhury, Ashley Gabriel, Ashwin Bharambe, Assaf Eisenman, Azadeh Yazdan, Beau James, Ben Maurer, Benjamin Leonhardi, Bernie Huang, Beth Loyd, Beto~De Paola, Bhargavi Paranjape, Bing Liu, Bo~Wu, Boyu Ni, Braden Hancock, Bram Wasti, Brandon Spence, Brani Stojkovic, Brian Gamido, Britt Montalvo, Carl
  Parker, Carly Burton, Catalina Mejia, Changhan Wang, Changkyu Kim, Chao Zhou, Chester Hu, Ching-Hsiang Chu, Chris Cai, Chris Tindal, Christoph Feichtenhofer, Damon Civin, Dana Beaty, Daniel Kreymer, Daniel Li, Danny Wyatt, David Adkins, David Xu, Davide Testuggine, Delia David, Devi Parikh, Diana Liskovich, Didem Foss, Dingkang Wang, Duc Le, Dustin Holland, Edward Dowling, Eissa Jamil, Elaine Montgomery, Eleonora Presani, Emily Hahn, Emily Wood, Erik Brinkman, Esteban Arcaute, Evan Dunbar, Evan Smothers, Fei Sun, Felix Kreuk, Feng Tian, Firat Ozgenel, Francesco Caggioni, Francisco Guzmán, Frank Kanayet, Frank Seide, Gabriela~Medina Florez, Gabriella Schwarz, Gada Badeer, Georgia Swee, Gil Halpern, Govind Thattai, Grant Herman, Grigory Sizov, Guangyi, Zhang, Guna Lakshminarayanan, Hamid Shojanazeri, Han Zou, Hannah Wang, Hanwen Zha, Haroun Habeeb, Harrison Rudolph, Helen Suk, Henry Aspegren, Hunter Goldman, Igor Molybog, Igor Tufanov, Irina-Elena Veliche, Itai Gat, Jake Weissman, James Geboski, James Kohli,
  Japhet Asher, Jean-Baptiste Gaya, Jeff Marcus, Jeff Tang, Jennifer Chan, Jenny Zhen, Jeremy Reizenstein, Jeremy Teboul, Jessica Zhong, Jian Jin, Jingyi Yang, Joe Cummings, Jon Carvill, Jon Shepard, Jonathan McPhie, Jonathan Torres, Josh Ginsburg, Junjie Wang, Kai Wu, Kam~Hou U, Karan Saxena, Karthik Prasad, Kartikay Khandelwal, Katayoun Zand, Kathy Matosich, Kaushik Veeraraghavan, Kelly Michelena, Keqian Li, Kun Huang, Kunal Chawla, Kushal Lakhotia, Kyle Huang, Lailin Chen, Lakshya Garg, Lavender A, Leandro Silva, Lee Bell, Lei Zhang, Liangpeng Guo, Licheng Yu, Liron Moshkovich, Luca Wehrstedt, Madian Khabsa, Manav Avalani, Manish Bhatt, Maria Tsimpoukelli, Martynas Mankus, Matan Hasson, Matthew Lennie, Matthias Reso, Maxim Groshev, Maxim Naumov, Maya Lathi, Meghan Keneally, Michael~L. Seltzer, Michal Valko, Michelle Restrepo, Mihir Patel, Mik Vyatskov, Mikayel Samvelyan, Mike Clark, Mike Macey, Mike Wang, Miquel~Jubert Hermoso, Mo~Metanat, Mohammad Rastegari, Munish Bansal, Nandhini Santhanam, Natascha
  Parks, Natasha White, Navyata Bawa, Nayan Singhal, Nick Egebo, Nicolas Usunier, Nikolay~Pavlovich Laptev, Ning Dong, Ning Zhang, Norman Cheng, Oleg Chernoguz, Olivia Hart, Omkar Salpekar, Ozlem Kalinli, Parkin Kent, Parth Parekh, Paul Saab, Pavan Balaji, Pedro Rittner, Philip Bontrager, Pierre Roux, Piotr Dollar, Polina Zvyagina, Prashant Ratanchandani, Pritish Yuvraj, Qian Liang, Rachad Alao, Rachel Rodriguez, Rafi Ayub, Raghotham Murthy, Raghu Nayani, Rahul Mitra, Raymond Li, Rebekkah Hogan, Robin Battey, Rocky Wang, Rohan Maheswari, Russ Howes, Ruty Rinott, Sai~Jayesh Bondu, Samyak Datta, Sara Chugh, Sara Hunt, Sargun Dhillon, Sasha Sidorov, Satadru Pan, Saurabh Verma, Seiji Yamamoto, Sharadh Ramaswamy, Shaun Lindsay, Shaun Lindsay, Sheng Feng, Shenghao Lin, Shengxin~Cindy Zha, Shiva Shankar, Shuqiang Zhang, Shuqiang Zhang, Sinong Wang, Sneha Agarwal, Soji Sajuyigbe, Soumith Chintala, Stephanie Max, Stephen Chen, Steve Kehoe, Steve Satterfield, Sudarshan Govindaprasad, Sumit Gupta, Sungmin Cho, Sunny
  Virk, Suraj Subramanian, Sy~Choudhury, Sydney Goldman, Tal Remez, Tamar Glaser, Tamara Best, Thilo Kohler, Thomas Robinson, Tianhe Li, Tianjun Zhang, Tim Matthews, Timothy Chou, Tzook Shaked, Varun Vontimitta, Victoria Ajayi, Victoria Montanez, Vijai Mohan, Vinay~Satish Kumar, Vishal Mangla, Vlad Ionescu, Vlad Poenaru, Vlad~Tiberiu Mihailescu, Vladimir Ivanov, Wei Li, Wenchen Wang, Wenwen Jiang, Wes Bouaziz, Will Constable, Xiaocheng Tang, Xiaofang Wang, Xiaojian Wu, Xiaolan Wang, Xide Xia, Xilun Wu, Xinbo Gao, Yanjun Chen, Ye~Hu, Ye~Jia, Ye~Qi, Yenda Li, Yilin Zhang, Ying Zhang, Yossi Adi, Youngjin Nam, Yu, Wang, Yuchen Hao, Yundi Qian, Yuzi He, Zach Rait, Zachary DeVito, Zef Rosnbrick, Zhaoduo Wen, Zhenyu Yang, and Zhiwei Zhao. 2024.
\newblock \href {https://arxiv.org/abs/2407.21783} {The llama 3 herd of models}.
\newblock \emph{Preprint}, arXiv:2407.21783.

\bibitem[{Eberhard(2019)}]{ethnologue_digitalsupport2024}
David Eberhard. 2019.
\newblock How to access and use the ethnologue, a curated repository of language information.
\newblock \url{https://sustainableheritagenetwork.org/}.
\newblock Accessed: 2024-07-29.

\bibitem[{Eberhard et~al.(2024)Eberhard, Simons, and Fennig}]{ethnologue2024}
David~M. Eberhard, Gary~F. Simons, and Charles~D. Fennig. 2024.
\newblock Ethnologue: Languages of the world.
\newblock \url{https://www.ethnologue.com}.
\newblock Accessed: 2024-07-29.

\bibitem[{Fishman(1972)}]{fishman1972language}
Joshua~A Fishman. 1972.
\newblock \emph{Language and Nationalism: Two Integrative Essays.}
\newblock ERIC.

\bibitem[{Fix and Hodges(1989)}]{60c19788-1128-3b5f-9275-2d63cc155832}
Evelyn Fix and J.~L. Hodges. 1989.
\newblock \href {http://www.jstor.org/stable/1403797} {Discriminatory analysis. nonparametric discrimination: Consistency properties}.
\newblock \emph{International Statistical Review / Revue Internationale de Statistique}, 57(3):238--247.

\bibitem[{for Standardization(2022)}]{iso15924}
International~Organization for Standardization. 2022.
\newblock Iso 15924: Information and documentation — codes for the representation of names of scripts.
\newblock \url{https://www.iso.org/standard/81905.html}.
\newblock Accessed: 2024-07-29.

\bibitem[{Friedman(2001)}]{friedman2001greedy}
Jerome~H Friedman. 2001.
\newblock Greedy function approximation: a gradient boosting machine.
\newblock \emph{Annals of statistics}, pages 1189--1232.

\bibitem[{Galton(1886)}]{55e7ba22-38fb-3d2b-9a2c-0e68080abfc3}
Francis Galton. 1886.
\newblock \href {http://www.jstor.org/stable/2841583} {Regression towards mediocrity in hereditary stature.}
\newblock \emph{The Journal of the Anthropological Institute of Great Britain and Ireland}, 15:246--263.

\bibitem[{Hammarström et~al.(2024)Hammarström, Forkel, Haspelmath, and Bank}]{glottolog2024}
Harald Hammarström, Robert Forkel, Martin Haspelmath, and Sebastian Bank. 2024.
\newblock Glottolog 5.0.
\newblock \url{https://glottolog.org}.
\newblock Accessed: 2024-07-29.

\bibitem[{Hoerl and Kennard(1970)}]{doi:10.1080/00401706.1970.10488634}
Arthur~E. Hoerl and Robert~W. Kennard. 1970.
\newblock \href {https://doi.org/10.1080/00401706.1970.10488634} {Ridge regression: Biased estimation for nonorthogonal problems}.
\newblock \emph{Technometrics}, 12(1):55--67.

\bibitem[{Imani et~al.(2023)Imani, Lin, Kargaran, Severini, Jalili~Sabet, Kassner, Ma, Schmid, Martins, Yvon, and Sch{\"u}tze}]{imanigooghari-etal-2023-glot500}
Ayyoob Imani, Peiqin Lin, Amir~Hossein Kargaran, Silvia Severini, Masoud Jalili~Sabet, Nora Kassner, Chunlan Ma, Helmut Schmid, Andr{\'e} Martins, Fran{\c{c}}ois Yvon, and Hinrich Sch{\"u}tze. 2023.
\newblock \href {https://doi.org/10.18653/v1/2023.acl-long.61} {Glot500: Scaling multilingual corpora and language models to 500 languages}.
\newblock In \emph{Proceedings of the 61st Annual Meeting of the Association for Computational Linguistics (Volume 1: Long Papers)}, pages 1082--1117, Toronto, Canada. Association for Computational Linguistics.

\bibitem[{International(2019)}]{ethnologue_vitality2024}
SIL International. 2019.
\newblock Mapping linguistic vitality and language endangerment.
\newblock \url{https://www.sil.org/resources/archives/78578}.
\newblock Accessed: 2024-07-29.

\bibitem[{Joshi et~al.(2020)Joshi, Santy, Budhiraja, Bali, and Choudhury}]{joshi-etal-2020-state}
Pratik Joshi, Sebastin Santy, Amar Budhiraja, Kalika Bali, and Monojit Choudhury. 2020.
\newblock \href {https://doi.org/10.18653/v1/2020.acl-main.560} {The state and fate of linguistic diversity and inclusion in the {NLP} world}.
\newblock In \emph{Proceedings of the 58th Annual Meeting of the Association for Computational Linguistics}, pages 6282--6293, Online. Association for Computational Linguistics.

\bibitem[{Kruskal(1964)}]{kruskal1964multidimensional}
Joseph~B Kruskal. 1964.
\newblock Multidimensional scaling by optimizing goodness of fit to a nonmetric hypothesis.
\newblock \emph{Psychometrika}, 29(1):1--27.

\bibitem[{Lample and Conneau(2019)}]{lample2019cross}
Guillaume Lample and Alexis Conneau. 2019.
\newblock Cross-lingual language model pretraining.
\newblock \emph{arXiv preprint arXiv:1901.07291}.

\bibitem[{Lin et~al.(2022)Lin, Mihaylov, Artetxe, Wang, Chen, Simig, Ott, Goyal, Bhosale, Du, Pasunuru, Shleifer, Koura, Chaudhary, O{'}Horo, Wang, Zettlemoyer, Kozareva, Diab, Stoyanov, and Li}]{lin-etal-2022-shot}
Xi~Victoria Lin, Todor Mihaylov, Mikel Artetxe, Tianlu Wang, Shuohui Chen, Daniel Simig, Myle Ott, Naman Goyal, Shruti Bhosale, Jingfei Du, Ramakanth Pasunuru, Sam Shleifer, Punit~Singh Koura, Vishrav Chaudhary, Brian O{'}Horo, Jeff Wang, Luke Zettlemoyer, Zornitsa Kozareva, Mona Diab, Veselin Stoyanov, and Xian Li. 2022.
\newblock \href {https://doi.org/10.18653/v1/2022.emnlp-main.616} {Few-shot learning with multilingual generative language models}.
\newblock In \emph{Proceedings of the 2022 Conference on Empirical Methods in Natural Language Processing}, pages 9019--9052, Abu Dhabi, United Arab Emirates. Association for Computational Linguistics.

\bibitem[{Lundberg and Lee(2017)}]{NIPS2017_7062}
Scott~M Lundberg and Su-In Lee. 2017.
\newblock \href {http://papers.nips.cc/paper/7062-a-unified-approach-to-interpreting-model-predictions.pdf} {A unified approach to interpreting model predictions}.
\newblock In I.~Guyon, U.~V. Luxburg, S.~Bengio, H.~Wallach, R.~Fergus, S.~Vishwanathan, and R.~Garnett, editors, \emph{Advances in Neural Information Processing Systems 30}, pages 4765--4774. Curran Associates, Inc.

\bibitem[{Muennighoff et~al.(2023)Muennighoff, Wang, Sutawika, Roberts, Biderman, Le~Scao, Bari, Shen, Yong, Schoelkopf, Tang, Radev, Aji, Almubarak, Albanie, Alyafeai, Webson, Raff, and Raffel}]{muennighoff-etal-2023-crosslingual}
Niklas Muennighoff, Thomas Wang, Lintang Sutawika, Adam Roberts, Stella Biderman, Teven Le~Scao, M~Saiful Bari, Sheng Shen, Zheng~Xin Yong, Hailey Schoelkopf, Xiangru Tang, Dragomir Radev, Alham~Fikri Aji, Khalid Almubarak, Samuel Albanie, Zaid Alyafeai, Albert Webson, Edward Raff, and Colin Raffel. 2023.
\newblock \href {https://doi.org/10.18653/v1/2023.acl-long.891} {Crosslingual generalization through multitask finetuning}.
\newblock In \emph{Proceedings of the 61st Annual Meeting of the Association for Computational Linguistics (Volume 1: Long Papers)}, pages 15991--16111, Toronto, Canada. Association for Computational Linguistics.

\bibitem[{Nezhad and Agrawal(2024)}]{nezhad2024exploringmazemultilingualmodeling}
Sina~Bagheri Nezhad and Ameeta Agrawal. 2024.
\newblock \href {https://arxiv.org/abs/2310.05404} {Exploring the maze of multilingual modeling}.
\newblock \emph{Preprint}, arXiv:2310.05404.

\bibitem[{NLLB et~al.(2022)}]{nllb2022flores}
NLLB et~al. 2022.
\newblock \href {https://arxiv.org/abs/2207.04672} {No language left behind: Scaling human-centered machine translation}.
\newblock \emph{arXiv preprint arXiv:2207.04672}.

\bibitem[{Philippy et~al.(2023)Philippy, Guo, and Haddadan}]{philippy-etal-2023-towards}
Fred Philippy, Siwen Guo, and Shohreh Haddadan. 2023.
\newblock \href {https://doi.org/10.18653/v1/2023.acl-long.323} {Towards a common understanding of contributing factors for cross-lingual transfer in multilingual language models: A review}.
\newblock In \emph{Proceedings of the 61st Annual Meeting of the Association for Computational Linguistics (Volume 1: Long Papers)}, pages 5877--5891, Toronto, Canada. Association for Computational Linguistics.

\bibitem[{Pires et~al.(2019)Pires, Schlinger, and Garrette}]{pires-etal-2019-multilingual}
Telmo Pires, Eva Schlinger, and Dan Garrette. 2019.
\newblock \href {https://doi.org/10.18653/v1/P19-1493} {How multilingual is multilingual {BERT}?}
\newblock In \emph{Proceedings of the 57th Annual Meeting of the Association for Computational Linguistics}, pages 4996--5001, Florence, Italy. Association for Computational Linguistics.

\bibitem[{Post(2018)}]{post-2018-call}
Matt Post. 2018.
\newblock \href {https://doi.org/10.18653/v1/W18-6319} {A call for clarity in reporting {BLEU} scores}.
\newblock In \emph{Proceedings of the Third Conference on Machine Translation: Research Papers}, pages 186--191, Brussels, Belgium. Association for Computational Linguistics.

\bibitem[{Quinlan(1986)}]{quinlan1986induction}
J.~Ross Quinlan. 1986.
\newblock Induction of decision trees.
\newblock \emph{Machine learning}, 1(1):81--106.

\bibitem[{Tibshirani(1996)}]{51791361-8fe2-38d5-959f-ae8d048b490d}
Robert Tibshirani. 1996.
\newblock \href {http://www.jstor.org/stable/2346178} {Regression shrinkage and selection via the lasso}.
\newblock \emph{Journal of the Royal Statistical Society. Series B (Methodological)}, 58(1):267--288.

\bibitem[{Vapnik et~al.(1995)Vapnik, Boser, Guyon, Vapnik, and Chervonenkis}]{Vapnik1995SupportVectorNetworks}
Vladimir Vapnik, Bernhard~E. Boser, Isabelle Guyon, Vladimir~N. Vapnik, and Alexey Chervonenkis. 1995.
\newblock \href {https://doi.org/10.1007/BF00994018} {Support-vector networks}.
\newblock \emph{Machine Learning}, 20:273--297.

\bibitem[{Wu and Dredze(2019)}]{wu-dredze-2019-beto}
Shijie Wu and Mark Dredze. 2019.
\newblock \href {https://doi.org/10.18653/v1/D19-1077} {Beto, bentz, becas: The surprising cross-lingual effectiveness of {BERT}}.
\newblock In \emph{Proceedings of the 2019 Conference on Empirical Methods in Natural Language Processing and the 9th International Joint Conference on Natural Language Processing (EMNLP-IJCNLP)}, pages 833--844, Hong Kong, China. Association for Computational Linguistics.

\bibitem[{Yang et~al.(2023)Yang, Li, Zhang, and Zong}]{yang2023bigtranslateaugmentinglargelanguage}
Wen Yang, Chong Li, Jiajun Zhang, and Chengqing Zong. 2023.
\newblock \href {https://arxiv.org/abs/2305.18098} {Bigtranslate: Augmenting large language models with multilingual translation capability over 100 languages}.
\newblock \emph{Preprint}, arXiv:2305.18098.

\bibitem[{Zhong et~al.(2024)Zhong, Cheng, Liu, Jiang, Wan, Chu, Murawaki, and Kurohashi}]{zhong2024englishcentricllmslanguagemultilingual}
Chengzhi Zhong, Fei Cheng, Qianying Liu, Junfeng Jiang, Zhen Wan, Chenhui Chu, Yugo Murawaki, and Sadao Kurohashi. 2024.
\newblock \href {https://arxiv.org/abs/2408.10811} {Beyond english-centric llms: What language do multilingual language models think in?}
\newblock \emph{Preprint}, arXiv:2408.10811.

\bibitem[{Zhu et~al.(2024)Zhu, Liu, Dong, Xu, Huang, Kong, Chen, and Li}]{zhu-etal-2024-multilingual}
Wenhao Zhu, Hongyi Liu, Qingxiu Dong, Jingjing Xu, Shujian Huang, Lingpeng Kong, Jiajun Chen, and Lei Li. 2024.
\newblock \href {https://doi.org/10.18653/v1/2024.findings-naacl.176} {Multilingual machine translation with large language models: Empirical results and analysis}.
\newblock In \emph{Findings of the Association for Computational Linguistics: NAACL 2024}, pages 2765--2781, Mexico City, Mexico. Association for Computational Linguistics.

\bibitem[{Zou and Hastie(2005)}]{10.1111/j.1467-9868.2005.00503.x}
Hui Zou and Trevor Hastie. 2005.
\newblock \href {https://doi.org/10.1111/j.1467-9868.2005.00503.x} {{Regularization and Variable Selection Via the Elastic Net}}.
\newblock \emph{Journal of the Royal Statistical Society Series B: Statistical Methodology}, 67(2):301--320.

\end{thebibliography}

\appendix
\newpage

\section{Appendix}
\label{sec:appendix}
The following tables present the performance metrics of various regression models evaluated for their effectiveness in predicting multilingual language model performance across different tasks and settings. Each table reports the R-squared values (indicating the proportion of variance explained by the model) along with Mean Squared Error (MSE) values, which provide insights into the model's accuracy.

Table \ref{tab:zero_shot_classification_results} shows the performance of different regression models when applied to zero-shot classification tasks using the Bloom, BloomZ, and XGLM models. The Random Forest and XGBoost models consistently achieve the highest R-squared values, indicating their strong ability to predict model performance accurately.

\begin{table*}[ht]
\centering
\caption{Performance of Regression Models for Zero-Shot Classification Tasks (R-squared with MSE in Parentheses)}
\begin{tabular}{lccc}
\toprule
\textbf{Model} & \textbf{Bloom} & \textbf{BloomZ} & \textbf{XGLM} \\
\midrule
Linear Regression & 0.354 (0.009) & 0.679 (0.003) & 0.627 (0.009) \\
Random Forest & \textbf{0.645 (0.005)} & \textbf{0.903 (0.001)} & 0.838 (0.004) \\
Decision Tree & 0.331 (0.009) & 0.842 (0.002) & 0.743 (0.006) \\
SVR & -0.018 (0.014) & 0.248 (0.007) & 0.033 (0.022) \\
Gradient Boosting & 0.623 (0.005) & 0.893 (0.001) & 0.807 (0.004) \\
XGBoost & 0.631 (0.005) & 0.866 (0.001) & \textbf{0.855 (0.003)} \\
K-Nearest Neighbors & -0.075 (0.015) & 0.369 (0.006) & -0.066 (0.025) \\
Lasso Regression & 0.001 (0.014) & 0.314 (0.007) & -0.017 (0.023) \\
Ridge Regression & 0.386 (0.009) & 0.695 (0.003) & 0.571 (0.010) \\
Elastic Net & 0.000 (0.014) & 0.313 (0.007) & -0.018 (0.023) \\
\bottomrule
\end{tabular}
\label{tab:zero_shot_classification_results}
\end{table*}

In two-shot classification tasks (Table \ref{tab:two_shot_classification_results}), the Gradient Boosting and XGBoost models perform well across the three multilingual models. 

\begin{table*}[ht]
\centering
\caption{Performance of Regression Models for Two-Shot Classification Tasks (R-squared with MSE in Parentheses)}
\begin{tabular}{lccc}
\toprule
\textbf{Model} & \textbf{Bloom} & \textbf{BloomZ} & \textbf{XGLM} \\
\midrule
Linear Regression & 0.593 (0.017) & 0.614 (0.012) & 0.658 (0.011) \\
Random Forest & 0.805 (0.008) & 0.676 (0.012) & 0.887 (0.004) \\
Decision Tree & 0.686 (0.013) & 0.380 (0.024) & 0.828 (0.005) \\
SVR & 0.248 (0.032) & 0.515 (0.018) & 0.013 (0.031) \\
Gradient Boosting & 0.800 (0.009) & \textbf{0.754 (0.009)} & 0.864 (0.004) \\
XGBoost & \textbf{0.847 (0.007)} & 0.693 (0.016) & \textbf{0.902 (0.003)} \\
K-Nearest Neighbors & 0.219 (0.034) & 0.420 (0.022) & -0.052 (0.033) \\
Lasso Regression & 0.278 (0.031) & 0.511 (0.019) & -0.061 (0.033) \\
Ridge Regression & 0.599 (0.017) & 0.686 (0.012) & 0.604 (0.012) \\
Elastic Net & 0.278 (0.031) & 0.511 (0.019) & -0.061 (0.033) \\
\bottomrule
\end{tabular}
\label{tab:two_shot_classification_results}
\end{table*}

Table \ref{tab:zero_shot_generation_results} highlights the performance of regression models for zero-shot generation tasks. Gradient Boosting and XGBoost models are particularly effective in this context, showing higher R-squared values and lower MSEs compared to other models, indicating their robustness in predicting performance without prior examples.

\begin{table*}[ht]
\centering
\caption{Performance of Regression Models for Zero-Shot Generation Tasks (R-squared with MSE in Parentheses)}
\begin{tabular}{lccc}
\toprule
\textbf{Model} & \textbf{Bloom} & \textbf{BloomZ} & \textbf{XGLM} \\
\midrule
Linear Regression & 0.402 (10.740) & 0.594 (186.307) & 0.457 (18.645) \\
Random Forest & 0.380 (11.135) & 0.890 (50.287) & 0.885 (3.932) \\
Decision Tree & -0.248 (22.426) & 0.751 (114.042) & 0.566 (14.894) \\
SVR & -0.002 (18.009) & 0.423 (264.669) & -0.092 (37.489) \\
Gradient Boosting & \textbf{0.553 (8.037)} & \textbf{0.918 (37.443)} & 0.876 (4.243) \\
XGBoost & 0.505 (8.889) & 0.894 (48.552) & \textbf{0.902 (3.365)} \\
K-Nearest Neighbors & 0.079 (16.549) & 0.639 (165.584) & -0.085 (37.239) \\
Lasso Regression & 0.194 (14.487) & 0.741 (118.974) & 0.121 (30.154) \\
Ridge Regression & 0.445 (9.970) & 0.652 (159.788) & 0.459 (18.557) \\
Elastic Net & 0.191 (14.537) & 0.731 (123.245) & 0.118 (30.257) \\
\bottomrule
\end{tabular}
\label{tab:zero_shot_generation_results}
\end{table*}

For two-shot generation tasks (Table \ref{tab:two_shot_generation_results}), the Gradient Boosting and XGBoost models continue to lead in performance. 

\begin{table*}[ht]
\centering
\caption{Performance of Regression Models for Two-Shot Generation Tasks (R-squared with MSE in Parentheses)}
\begin{tabular}{lccc}
\toprule
\textbf{Model} & \textbf{Bloom} & \textbf{BloomZ} & \textbf{XGLM} \\
\midrule
Linear Regression & 0.574 (20.081) & 0.819 (68.265) & 0.448 (8.193) \\
Random Forest & 0.820 (8.481) & 0.924 (28.792) & 0.765 (3.485) \\
Decision Tree & 0.651 (16.454) & 0.899 (38.059) & 0.571 (6.371) \\
SVR & -0.043 (49.111) & 0.230 (290.308) & -0.120 (16.633) \\
Gradient Boosting & 0.844 (7.340) & \textbf{0.950 (18.687)} & \textbf{0.801 (2.950)} \\
XGBoost & \textbf{0.866 (6.322)} & 0.884 (43.924) & 0.636 (5.409) \\
K-Nearest Neighbors & 0.041 (45.137) & 0.437 (212.228) & -0.062 (15.782) \\
Lasso Regression & 0.141 (40.439) & 0.793 (78.051) & 0.080 (13.666) \\
Ridge Regression & 0.584 (19.606) & 0.826 (65.626) & 0.440 (8.313) \\
Elastic Net & 0.141 (40.439) & 0.757 (91.790) & 0.100 (13.376) \\
\bottomrule
\end{tabular}
\label{tab:two_shot_generation_results}
\end{table*}

These tables underscores the advantage of these ensemble methods in capturing complex feature interactions in multilingual language models.

\end{document}